\documentclass[conference]{IEEEtran}
\IEEEoverridecommandlockouts
\usepackage{cite}
\usepackage{amsmath,amssymb,amsfonts}
\usepackage{algorithm}
\usepackage{algorithmic}
\usepackage{graphicx}
\usepackage{textcomp}
\usepackage{xcolor}
\usepackage[normalem]{ulem}
\def\BibTeX{{\rm B\kern-.05em{\sc i\kern-.025em b}\kern-.08em
    T\kern-.1667em\lower.7ex\hbox{E}\kern-.125emX}}
\makeatletter
\newcommand{\linebreakand}{
  \end{@IEEEauthorhalign}
  \hfill\mbox{}\par
  \mbox{}\hfill\begin{@IEEEauthorhalign}
}
\makeatother
\begin{document}

\title{CGAR: Critic Guided Action Redistribution in Reinforcement Leaning}

\author{\IEEEauthorblockN{Tairan Huang}
\IEEEauthorblockA{\textit{SCSE} \\
\textit{Beihang University}\\
Beijing, China \\
trhuang@buaa.edu.cn}
\and
\IEEEauthorblockN{Xu Li}
\IEEEauthorblockA{\textit{Cognitive Computing Lab} \\
\textit{Baidu Research}\\
Beijing, China \\
lixu13@baidu.com}
\and
\IEEEauthorblockN{Hao Li}
\IEEEauthorblockA{\textit{ECE} \\
\textit{Peking University}\\
Beijing, China \\
2101212812@pku.edu.cn}
\linebreakand
\IEEEauthorblockN{Mingming Sun}
\IEEEauthorblockA{\textit{Cognitive Computing Lab} \\
\textit{Baidu Research}\\
Beijing, China \\
sunmingming01@baidu.com}
\and
\IEEEauthorblockN{Ping Li}
\IEEEauthorblockA{\textit{Cognitive Computing Lab} \\
\textit{Baidu Research}\\
Seattle, USA \\
liping11@baidu.com}
}

\maketitle

\begin{abstract}
Training a game-playing reinforcement learning agent requires multiple interactions with the environment.
Ignorant random exploration may cause a waste of time and resources.
It's essential to alleviate such waste.
As discussed in this paper, under the settings of the off-policy actor critic algorithms, 
we demonstrate that the critic can bring more expected discounted rewards than or at least equal to the actor. 
Thus, the Q value predicted by the critic is a better signal to redistribute the action originally sampled from the policy distribution predicted by the actor.
This paper introduces the novel Critic Guided Action Redistribution (CGAR) algorithm
and tests it on the OpenAI MuJoCo tasks.
The experimental results demonstrate that our method improves the sample efficiency and achieves state-of-the-art performance. 
Our code can be found at https://github.com/tairanhuang/CGAR.
\end{abstract}


\section{Introduction}
In recent years, reinforcement learning has been widely used in games and has made excellent progress in Atari, StarCraft, Dota2, Honor of Kings, and other games
~\cite{DBLP:conf/nips/YeCZCYLCLQYYSWS20,DBLP:conf/nips/BrownBLG20}.
However, training a reinforcement learning model is time-consuming due to the massive interactions between the learning and the environment
~\cite{sutton2011reinforcement,mnih2015human,DBLP:conf/cig/MalloySK0RT21}. 
Also, interactions with naive exploration strategies slow down the model's learning speed 
and waste resources during model training
~\cite{DBLP:conf/cig/LiangZTYZ21,DBLP:conf/cig/ButtnerM21}.
These deficiencies restrict the applications of reinforcement learning in games.  

Off-policy algorithms store the experience in the buffer and reuse them to reduce interaction costs. 
Actor critic algorithms use an actor to select actions and use a critic to estimate the value function to reduce the variance of policy gradient and accelerate the convergence.
Deep deterministic policy gradient (DDPG)~\cite{DBLP:journals/corr/LillicrapHPHETS15} and Soft Actor-Critic (SAC) algorithm~\cite{DBLP:conf/icml/HaarnojaZAL18} combine the advantages of off-policy algorithms and actor critic algorithms. DDPG learns a critic and an actor at the same time. It uses the Bellman function to optimize the critic and uses the critic to optimize the actor. SAC~\cite{DBLP:conf/icml/HaarnojaZAL18} maximizes both expected return and entropy~\cite{DBLP:conf/icml/HaarnojaTAL17,DBLP:conf/aaai/ZiebartMBD08} to balance exploration and exploitation. The actor of SAC predicts the action distribution and then samples action in the training stage and uses the mean value of the distribution as the action in the evaluation stage. It further improves the sample efficiency and stability of reinforcement learning.

Following the same purpose, we propose a novel Critic Guided Action Redistribution (CGAR) mechanism and show that SAC with CGAR achieves state-of-the-art performance.
In this paper, we first give a theoretical analysis of actor critic algorithms 
and demonstrate that the critic can bring more expected discounted rewards than or at least equal to the actor 
in off-policy actor critic algorithms. 
To utilize such advancement of critic, we redistribute the action distribution predicted by the actor through the Q value predicted by the critic. 

\begin{figure*}[h]
\vspace{-0.6in}
\begin{center}
\includegraphics[width=5.0in]
{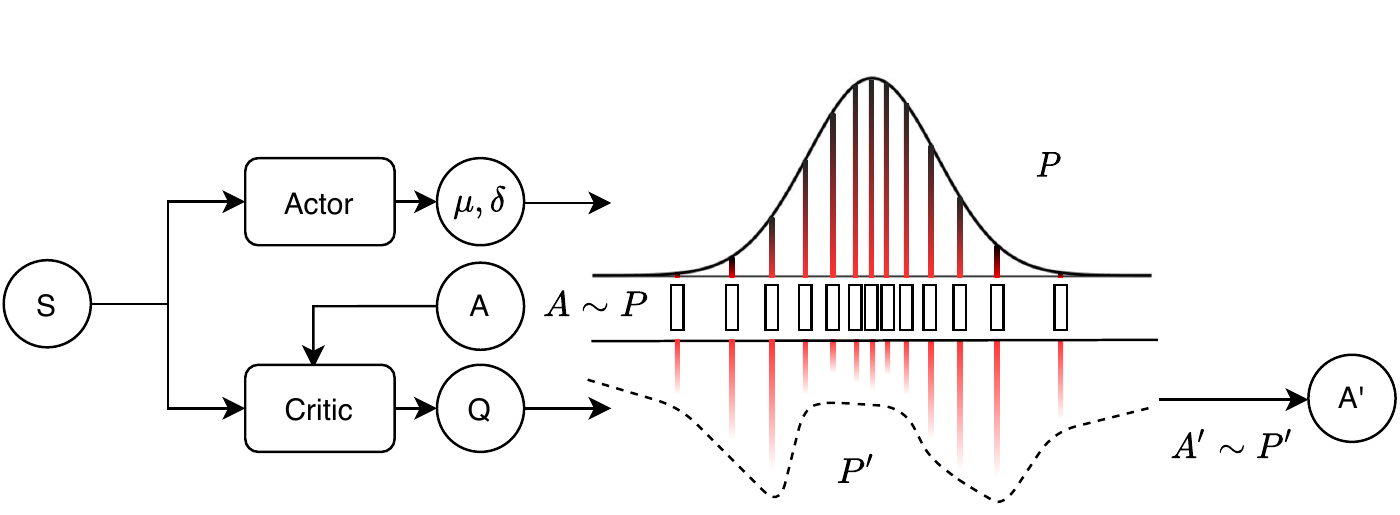} 
\vspace{-0.1in}
\end{center}
\caption{CGAR: Critic Guided Action Redistribution in Reinforcement Leaning} 
\label{ddpg} 
\vspace{-0.1in}
\end{figure*}

In the proposed CGAR model, after getting an action distribution predicted by the actor, 
we sample $K$ actions from that action distribution. 
Then, the critic model predicts the Q values for these $K$ actions conditioned on the state. 
We reset the selection probability of these actions based on their corresponding Q values.
Since the critic model is optimized to predict the expected return, with the distribution positively correlated to the Q value, the algorithm tends to select the action with a higher expected return.
We deploy our method on SAC and conduct experiments on the OpenAI MuJoCo tasks~\cite{DBLP:journals/corr/abs-1801-00690}. The experimental results demonstrate that our approach is effective and achieves state-of-the-art performance. 
Here we summarize our main contributions:
\begin{itemize}
\item We demonstrate that the critic can bring more expected discounted rewards than or at least equal to the actor in the off-policy actor critic algorithm.
\item We propose a novel Critic Guided Action Redistribution (CGAR), which uses the Q value predicted by the critic to resample action from the action distribution predicted by the actor.
\item We apply our method to SAC and achieve state-of-the-art performance on OpenAI MuJoCo tasks.
\end{itemize}

\section{Algorithm}

\subsection{Motivation}\label{sec_2_2}

Under the classic setting of the off-policy actor critic learning procedure, at each interaction with the environment (environment step), the agent collects data from the environment under the policy network with the action distribution of ${\pi_{\phi_a(s)}}$. Then, given the updated data buffer $\mathcal D$, the critic is optimized to estimate the future reward. Then, the actor is updated to maximize the estimated expected future reward given the currently learned critic. 
The procedure forms a dependency circle and loops at each environment step during training.

At a certain environment step, suppose that the critic ${Q_{\phi_c}}$ is optimized to ${Q_{\phi_c^i}}$ after $i$ gradient step under the currently collected data $\mathcal D$. The loss function to optimize actor is generally written as,
\begin{equation}
J_{\pi}(\phi_a) = \mathbb{E}_{s_t \sim \mathcal D} \mathbb{E}_{a_t \sim \mathcal \pi_{\phi_a}(a_t|s_t)} (-Q_{\phi_c^{i}}(s_t,a_t)).
\label{eq5}
\end{equation}
When the loss function is minimized, the action with the highest probability predicted by the actor is the same as the action leading to the highest ${Q_{\phi_c^i}}$ predicted by the critic. Then the performance of the actor and the critic is the same. However, most of the time, the actor is less optimized to the optimal distribution of action given the current critic, which leads to poor performance compared with the current critic.

This learning procedure is similar to the knowledge distillation or teacher-student method 
\cite{44873}.
In our situation, we use the Q value output by the critic to calculate the loss function to train the actor. Hence, critic and actor correspond to teacher and student, respectively.
The performance gap between the student and teacher has been demonstrated in many previous works, and the gap exists even though the student network has the same size as the teacher network 
~\cite{mirzadeh2020improved,cho2019efficacy,deng2020can}. 
In Section \ref{sec:em_mot}, we also give an empirical demonstration of the motivation.
As the performance of the critic is better than or at least equal to the actor in each environment step, 
we can expect that the critic can bring more expected discounted rewards than or at least equal to the actor during the RL training procedure.
This paper proposes a CGAR algorithm to ameliorate the performance gap between the actor and critic.
\subsection{Empirical Demonstration of the Motivation}\label{sec:em_mot}
\begin{figure}[h] 
\centering
\includegraphics[width=0.45\textwidth]{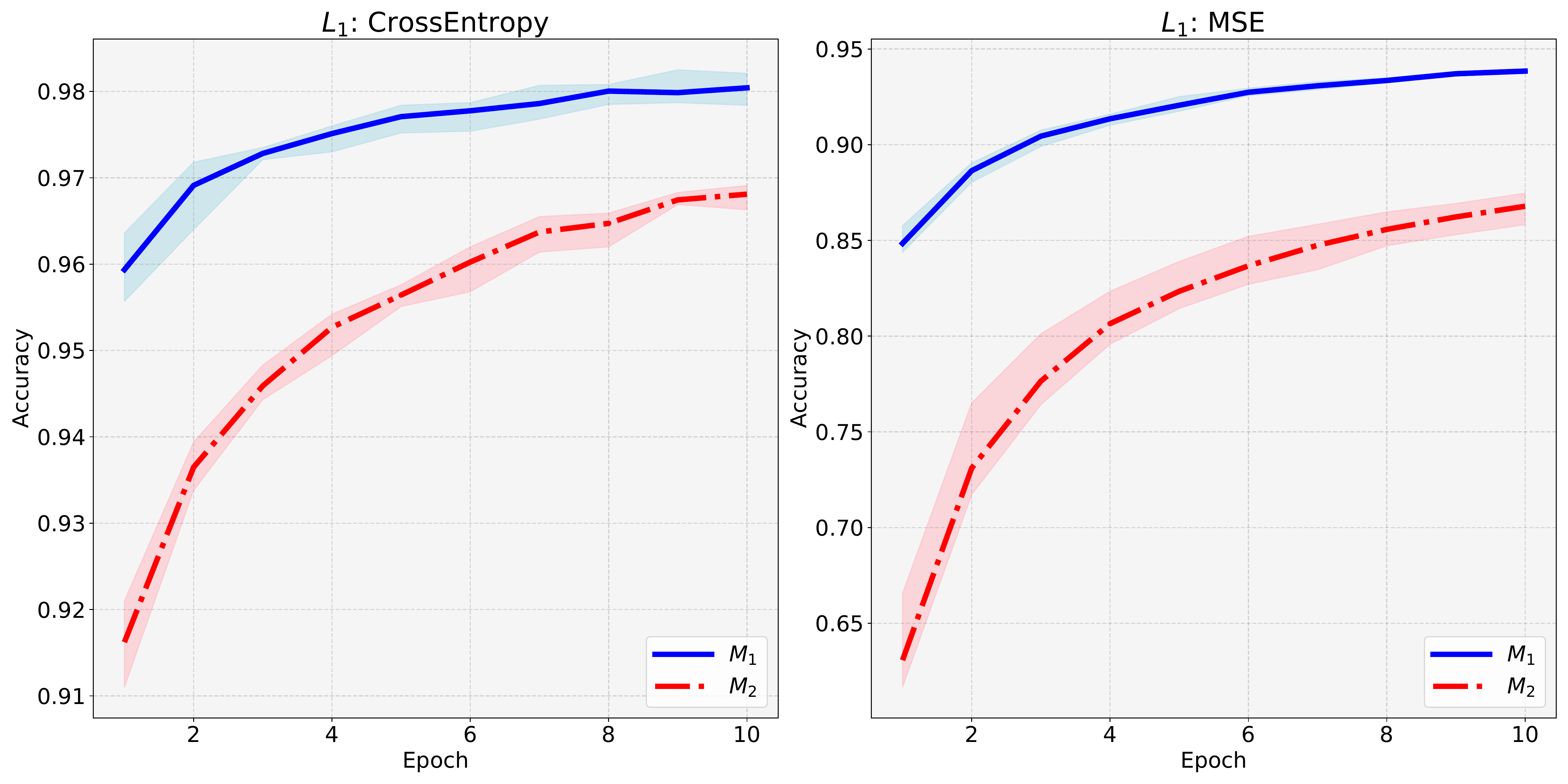} 
\caption{Comparison results of $M_1$ and $M_2$.} 
\label{fig1} 
\vspace{-0.1in}
\end{figure}

We conduct experiments under supervised learning settings to empirically demonstrate the above motivation. In detail, We use MNIST~\cite{lecun-mnisthandwrittendigit-2010} as a dataset and maintain two identical multilayer perceptron models, $M_1$ and $M_2$. 
We let $M_1$ fit the dataset and let $M_2$ fit $M_1$. Given input $x$ and its label $y$, the loss function of $M_1$ is: ${\mathcal{L}_1=CrossEntropy(M_1(x),y)}$, and the loss function of $M_2$ is: ${\mathcal{L}_2=CrossEntropy(M_2(x),M_1(x))}$.
Under this setting, $M_1$ is the signal provided to $M_2$, and $y$ is the signal provided to $M_1$. Besides, $M_1$ and $M_2$ are updated one after another iteratively. 
This setting is quite similar to the off-policy actor critic algorithm in Section ~\ref{sec_2_2}, where the reward is provided to calculate the critic's target, and the Q is the signal provided to the actor.
To simulate the loss function of critic, we further design another loss function, ${\mathcal{L}_1=MSE(M_1(x),\bar{y})}$, where $\bar{y}$ is the one-hot representation of $y$, and ${\mathcal{L}_2}$ remains unchanged.
We test the evaluation accuracy of $M_1$ and $M_2$ every epoch over five seeds and report the results in Fig.~\ref{fig1}.
We can see that under both loss functions, $M_2$ learns more slowly than $M_1$, demonstrating our motivation's correctness.

\subsection{Critic Guided Action Redistribution}
We propose our Critic Guided Action Redistribution (CGAR) based on the above motivation and empirical demonstration.
In actor critic algorithms, given state ${s_t}$ at environment step $t$, the actor ${\pi_{\phi_a}}$ predicts the action distribution ${\mathcal{P}_t}$. After that, it samples action ${a_t}$ from ${\mathcal{P}_t}$ to interact with the environment. 
\begin{equation}
a_t \sim \mathcal{P}_t = \pi_{\phi_a}(s_t).
\label{eq8}
\end{equation}
In our algorithm,
we first sample $K$ actions ${\{a_t^0, a_t^1,...,a_t^{K-1}\}}$ from $\mathcal{P}_t$. These actions construct the actions set ${\mathbb{L}_a^t}$. 
\begin{equation}
\mathbb{L}_a^t := \{a_t^i\sim \mathcal{P}_t|i \in [0, K-1]\}.
\label{eq10}
\end{equation}
Then we use the critic to predict the Q value for every action in ${\mathbb{L}_a^t}$. The Q value ${Q_t^i}$ is calculated on every action ${a_t^i}$ in ${\mathbb{L}_a^t}$ conditioned with state $s_t$. These Q values ${\{Q_t^0, Q_t^1,...,Q_t^{K-1}\}}$ construct the Q value set ${\mathbb{L}_Q^t}$.  
\begin{equation}
\mathbb{L}_Q^t := \{Q_t^i=Q_{\phi_c}(s_t,a_t^i)|i \in [0, K-1]\}.
\label{eq11}
\end{equation}
We use the Q value ${Q_t^i}$ to calculate the probability ${p_t^i}$ of ${a_t^i}$. The probability set results from the Q value set calculated by the Softmax function. In this way, we could get the action probability distribution $\mathcal{P}_t'$.
\begin{equation}
\begin{aligned}
\mathcal{P}_t' &= \text{Softmax}(\mathbb{L}_Q^t) \\
&= \{\displaystyle\frac{e^{Q_t^i}}{\sum_{j=0}^{K-1}e^{Q_t^j}}|i \in [0, K-1]\}.
\label{eq12}
\end{aligned}
\end{equation}
We use the Softmax function to make actions with large Q values more likely to be sampled while maintaining a certain degree of exploration. Finally, we select the action ${a_t}$ from the new distribution. Fig.~\ref{ddpg} is our model's diagram.
\begin{equation}
a_t \sim \mathcal{P}'_t.
\label{eq14}
\end{equation}
\subsection{Implementation}
We apply CGAR to SAC. SAC is an off-policy algorithm based on the Maximum Entropy Principle.
Its optimization goal is simultaneously maximizing both the expected return and the entropy.
It learns a policy ${\pi_{\phi_a}}$, a Q value function ${Q_{\phi_c}}$, and a temperature coefficient ${\alpha}$, with parameters ${\phi_a}$, ${\phi_c=(\phi_c^1,\phi_c^2)}$, and ${\alpha}$ separately. 
The loss functions of SAC are defined below, which are introduced in the SAC paper.
\begin{equation}
\begin{aligned}
J_Q(\phi_c) = & \mathbb{E}_{s_t,a_t \sim \mathcal D} \frac{1}{2}(Q_{\phi_c}(s_t,a_t)
\\
-&(r_t+\gamma\mathbb{E}_{s_{t+1} \sim p} (V_{\bar{\phi}_c}(s_{t+1}))))^2,
\end{aligned}
\label{eq15}
\end{equation}
\begin{equation}
\begin{aligned}
V_{\bar{\phi}_c}(s_t) =& \mathbb{E}_{a_t \sim \pi_{\phi_a}(a_t|s_t)} Q_{\bar{\phi}_c}(s_{t},a_t) 
    \\-& \alpha \log \pi_{\phi_a}(a_{t}|s_{t}),
\end{aligned}
\label{eq16}
\end{equation}
\begin{equation}
\begin{aligned}
J_{\pi}(\phi_a) = & \mathbb{E}_{s_t \sim \mathcal D} \mathbb{E}_{a_t \sim \mathcal \pi_{\phi_a}(a_t|s_t)} 
\\&(\alpha \log \pi_{\phi_a}(a_t|s_t) - Q_{\phi_c}(s_t,a_t)).
\end{aligned}
\label{eq17}
\end{equation}
\begin{equation}
J_{\alpha}(\alpha) =  \mathbb{E}_{a_t \sim \mathcal \pi_{\phi_a}(a_t|s_t)} - \alpha \log \pi_{\phi_a}(a_t|s_t)  - \alpha \mathcal H,
\label{eq18}
\end{equation}
Note that different from Eq.~\eqref{eq5}, there's an entropy term in Eq.~\eqref{eq17}, which doesn't affect the demonstration in Section~\ref{sec_2_2}.
Our complete algorithm is shown in Algorithm~\ref{alg1}. Our unique operations are marked with red. And the deleted operations in SAC are marked with blue and strikethrough.

\begin{algorithm}[H]
\caption{CGAR applied to SAC}
\begin{algorithmic}[1]
\STATE {\bfseries Input}: \textsc{env}
\STATE {\bfseries Output}: $\phi_c$, $\phi_a$ and $\alpha$
\STATE Initialize parameters $\phi_c$, $\phi_a$ and $\alpha$
\STATE Set $\bar{\phi}_c=\phi_c$, $\mathcal D = \emptyset$

\FOR{$k=1,\dots,N_{\rm init}$}
\STATE Sample random action: ${a_t \sim \pi_{random}(s_t)}$ 
\STATE Execute action: ${s_{t+1}, r_{t+1}, \rm done \sim \textsc{env}(a_t)}$
\STATE Collect data: ${\mathcal D \leftarrow \mathcal D \cup\{s_t,a_t,s_{t+1},r_{t+1}\}}$
\ENDFOR
\FOR{$k=1,\dots,N_{\rm train}$}  
\STATE Predict action distribution: ${\mathcal{P}_t = \pi(s_t)}$
\STATE \sout{\color{blue}{Sample action: ${a_t \sim \mathcal{P}_t}$}}
\STATE {\color{red}{Sample ${K}$ actions from ${\mathcal{P}_t}$: \\
${\mathbb{L}_a^t = \{a_t^i\sim \mathcal{P}_t|i \in [0, K-1]\}}$}}
\STATE {\color{red}{Calculate the Q value set:\\ ${\mathbb{L}_Q^t = \{Q_t^i=Q_{\phi_c}(s_t,a_t^i)|i \in [0, K-1]\}}$}}
\STATE {\color{red}{Get the new distribution: ${\mathcal{P}'_t = \text{Softmax}(\mathbb{L}_Q^t)}$}}
\STATE {\color{red}{Sample action: ${a_t \sim \mathcal{P}'_t}$}}
\STATE Execute action: ${s_{t+1}, r_{t+1}, \rm done \sim \textsc{env}(a_t)}$
\STATE Collect data: ${\mathcal D \leftarrow \mathcal D \cup\{s_t,a_t,s_{t+1},r_{t+1}\}}$
\STATE Train critic: ${\phi_c \leftarrow \phi_c - \lambda_Q\nabla_{\phi_c} J_Q(\phi_c)}$
\STATE Train actor: ${\phi_a \leftarrow \phi_a - \lambda_{\pi}\nabla_{\phi_a} J_{\pi}(\phi_a)}$
\STATE Train alpha: ${\alpha \leftarrow \alpha-\lambda_{\alpha}\nabla_{\alpha} J_{\alpha}(\alpha)}$
\STATE Update the target critic: ${\bar\phi_c \leftarrow \tau\phi_c + (1-\tau)\bar\phi_c}$
\ENDFOR
\end{algorithmic}
\label{alg1}
\end{algorithm}

\section{Experiment} \label{sec_3}

\subsection{Implementation Details}
We implement CGAR on the SAC algorithm noted as CGAR-SAC. 
The implementation of the SAC comes from ~\cite{pytorch_sac}.
We evaluate our method on OpenAI MuJoCo tasks, including tasks such as standing, walking, and running. The state of the agent consists of parameters such as positions and velocities. Action is a real-valued vector that represents the control of the agent's joints. The purpose of model learning is to maximize the expected discounted rewards.
We test the agent performance every 10,000 environment steps. We compute the mean episode returns an agent obtains over ten episodes for every evaluation.
All results are over five different seeds, and we keep the minimum, maximum, and mean values over these seeds. 
\subsection{Comparison between CGAR-SAC with SAC}
\begin{figure}[h]
\begin{center}
\vspace{-0.1in}
\includegraphics[width=3.2in]
{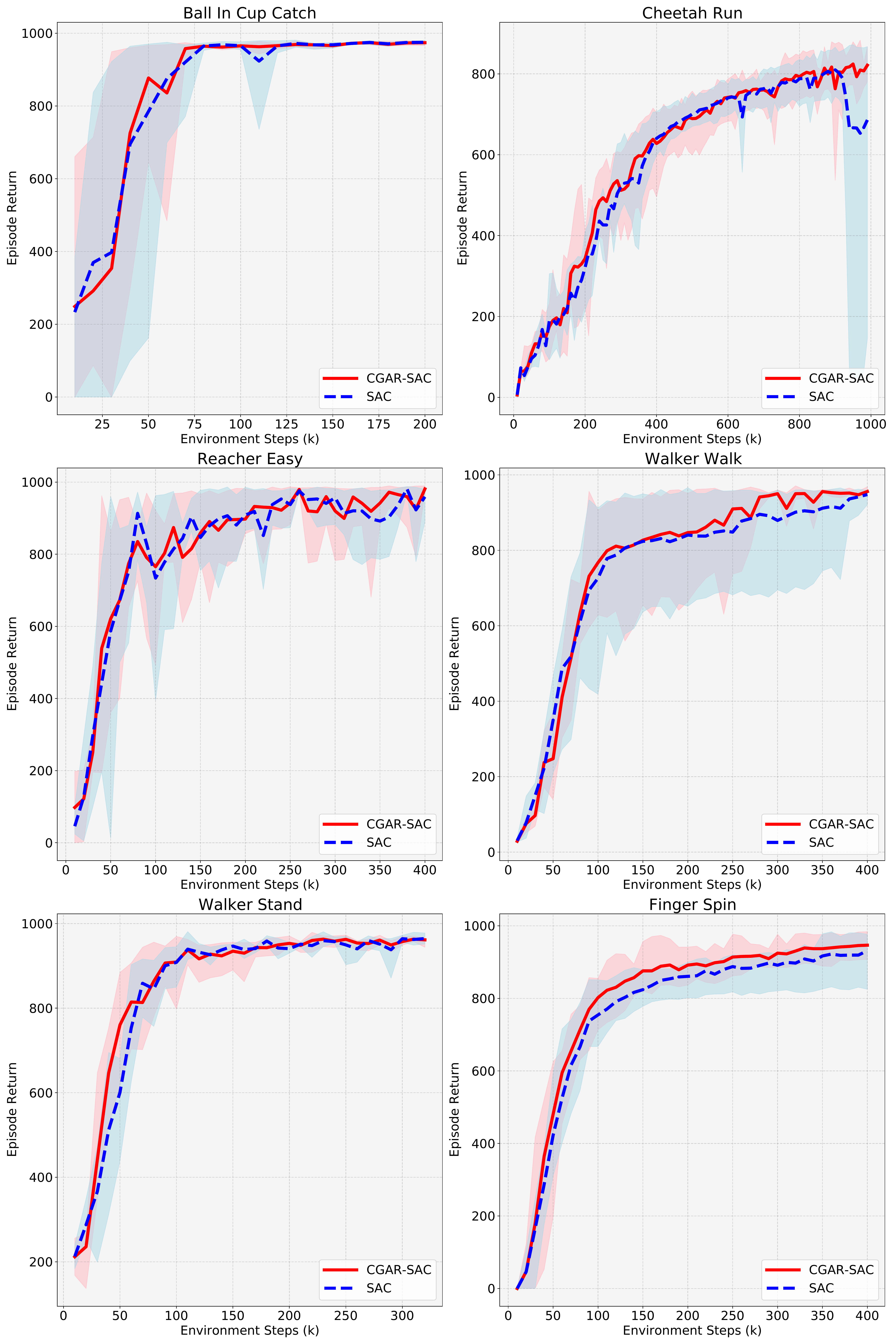} 
\vspace{-0.1in}
\end{center}
\caption{Comparison results between CGAR-SAC and SAC.} 
\label{fig2} 
\vspace{-0.1in}
\end{figure}

\begin{table*}
\caption{Mean value of the average return for the whole training process}
\begin{center}
\begin{tabular}{|l|llllll|}
\hline
Task & BCC & 
CR & RE & WW &WS &FS\\ \hline
CGAR-SAC & ${\bf{843_{\pm73}}}$ & 
${\bf{593_{\pm51}}}$ & 
${\bf{821_{\pm81}}}$ & 
${\bf{761_{\pm71}}}$ & 
${{{\bf{857}}_{\pm32}}}$ & 
${\bf{793_{\pm47}}}$\\
SAC & ${{841_{\pm89}}}$ 
& ${{574_{\pm59}}}$
& ${{817_{\pm82}}}$ 
& ${{746_{\pm110}}}$ 
& ${{846_{\pm\bf{29}}}}$ 
& ${{760_{\pm49}}}$\\
\hline
\end{tabular}
\end{center}
\label{tab1}
\vspace{-0.2in}
\end{table*}
We compare the performance of CGAR-SAC with SAC.
We counted the mean, maximum and minimum values of the average return obtained overall seeds during each evaluation and plotted them in Fig.~\ref{fig2}. The curve represents the average return, and the shading represents the range between the maximum and minimum. 
From Fig.~\ref{fig2} we can see that in most tasks, CGAR-SAC converges faster and achieves better final performance than SAC, especially Cheetah Run, Walker Walk, and Finger Spin. And CGAR-SAC is not weaker than SAC in other tasks. 
We also counted the average return
obtained overall seeds during each evaluation and calculated their mean over the entire training process. In Table~\ref{tab1}, we use the initials of the first letter of every environment as the table title, and the full names can be found in Fig.~\ref{fig2}. The value of each item in the table represents the mean value of the average return, and the subscript denotes the standard deviation.
From Table~\ref{tab1}, we can see that the mean value of the average return of our method during the training process is higher than SAC in every task, and the standard deviation is lower than SAC in most tasks. 
We can conclude that CGAR improves the sample efficiency of SAC.

\section{Conclusion}\label{sec_6}
This paper proposes a novel action redistribution algorithm, Critic Guided Action Redistribution for game playing. We demonstrate that the critic 
can bring more expected discounted rewards than or at least equal to the actor
in the off-policy actor critic algorithm.
Based on the demonstration, we use the Q value predicted by the critic to redistribute the actions probability distribution generated by the actor.
Then we sample actions from the new distribution to interact with the environment. We implement our algorithm on SAC and test it on the OpenAI MuJoCo tasks. The experimental results demonstrate that our method improves the sample efficiency and achieves state-of-the-art performance. 
Future research can be done by applying CGAR to other games or analyzing the distribution of Q value on the multimodal distribution.

\bibliography{cog_2022}
\bibliographystyle{IEEEtran}

\end{document}